\documentclass[runningheads]{llncs}
\usepackage[T1]{fontenc}
\usepackage{graphicx}
\usepackage[hidelinks]{hyperref}
\usepackage{tabularx}
\usepackage{xcolor}

\begin{document}
\title{A Benchmark Suite of Reddit-Derived Datasets for Mental Health Detection}
%
%
\author{
Khalid Hasan \and
Jamil Saquer
}
\authorrunning{K. Hasan et al.}
%
\institute{
Missouri State University, Springfield, MO, USA\\
\email{\{kh597s,jamilsaquer\}@missouristate.edu}}
\maketitle              
%



\begin{abstract}

The growing availability of online support groups has opened up new windows to study mental health through natural language processing (NLP). However, it is hindered by a lack of high-quality, well-validated datasets. Existing studies have a tendency to build task-specific corpora without collecting them into widely available resources, and this makes reproducibility as well as cross-task comparison challenging. 
In this paper, we present a uniform benchmark set of four Reddit-based datasets for disjoint but complementary tasks: (i) detection of suicidal ideation, (ii) binary general mental disorder detection, (iii) bipolar disorder detection, and (iv) multi-class mental disorder classification. All datasets were established upon diligent linguistic inspection, well-defined annotation guidelines, and human-judgmental verification. Inter-annotator agreement metrics always exceeded the baseline agreement score of 0.8, ensuring the labels' trustworthiness.
Previous work's evidence of performance on both transformer and contextualized recurrent models demonstrates that these models receive excellent performances on tasks (F1 $\approx$ 93–99\%), further validating the usefulness of the datasets. By combining these resources, we establish a unifying foundation for reproducible mental health NLP studies with the ability to carry out cross-task benchmarking, multi-task learning, and fair model comparison. The presented benchmark suite provides the research community with an easy-to-access and varied resource for advancing computational approaches toward mental health research.


\end{abstract}

\section{Introduction}

Mental illness disorders such as depression, anxiety, bipolar disorder, and post-traumatic stress disorder (PTSD) are a worldwide issue of concern to hundreds of millions of individuals worldwide~\cite{who@mentalhealth}. Though online support groups for mental health are ubiquitous, particularly on platforms such as Reddit, automated mental health language detection and analysis remain hindered by a fundamental bottleneck: the inability to leverage known, standardized datasets. Model building is isolated without these datasets, and reproducibility between studies is poor.

Recent improvements in natural language processing (NLP), particularly transformer-based models~\cite{devlin2019bert}~\cite{liu2019roberta}, have shown promise in detecting mental health markers from web content. Their performance, however, largely relies upon the quality and consistency of the underlying dataset. While prior work has proposed tasks such as suicidal ideation identification, disorder-specific identification, and multi-class classification, no end-to-end benchmark set is available to facilitate fair comparison, reproducibility, and generalizability across tasks.

To bridge this gap, we introduce a universal benchmark dataset from Reddit mental illness communities. The benchmark is a consolidation of four datasets that we originally introduced in task-focused experiments: Suicidal Ideation Detection~\cite{hasan2024suicidal}, General Mental Disorder Detection~\cite{hasan2025generalmental}, Bipolar Disorder Detection~\cite {hasan2025bipolar}, and Multi-Class Mental Disorder Classification~\cite{hasan2025multiMental}. The linguistic analysis process, annotation guidelines, and judgmental validation carefully crafted the datasets to ensure high reliability and replicability. In earlier research, we demonstrated the utility of these datasets separately by training and testing models such as RoBERTa, BERT, and LSTM variants. In this paper, however, we do not seek to present new models but to combine these datasets into a benchmark suite that highlights their overall consistency, linguistic diversity, and utility to future study. We make the datasets available for other researchers at Zenodo~\cite{khalid2025mentaldataset}.

We make three contributions:

\begin{enumerate}
    \item \textbf{Dataset Resource Consolidation:} We introduce four empirically supported Reddit-based datasets as a standard benchmark set, where all the datasets contain one task in mental health detection.

    \item \textbf{Empirical and Human Validation:} We characterize dataset strength through linguistic analysis, annotation guidelines, inter-annotator agreement, and reported benchmark performance by prior work.

    \item \textbf{Benchmark Potential:} We stake a claim for future use of these datasets for cross-task comparison, multi-task learning, and standardized benchmarking, and thus making them a foundation for reproducible mental health NLP research.
\end{enumerate}

With this benchmark set, we hope to provide the research community with a reproducible, reliable, and heterogeneous foundation for equitable model comparison, facilitating multi-task methods, and simplifying further development of NLP applications to mental health.

\section{Related Work}

The intersection of natural language processing (NLP) and mental health research has grown exponentially over the last few years, driven by large amounts of user-posted text available on websites like Reddit, Twitter (now X), and dedicated support sites. Previous research has established that linguistic markers are effective predictors of psychological distress, suicidality, and some disorders such as depression, anxiety, or bipolar disorders~\cite{coppersmith2015adhd}~\cite{de2013predicting}. These studies established the feasibility of social media data usage in mental health detection, but were often plagued by small sample sizes or task-specific data.

Several datasets have since been made available to examine online discussion on mental disorders. Initial datasets such as the CLPsych shared tasks~\cite{milne2016clpsych} focused on depression and suicide risk prediction based on tweets, while others, such as DAIC-WOZ~\cite{gratch2014distress}, explored multi-modal signals from clinical interviews. More recently, Reddit has been a treasure trove for large datasets~\cite{zirikly2019clpsych} with its topic-centered specialist communities and unstructured, anonymous posts. These projects tend to be strongly specialized in a single activity or application, however, and hence are of lesser general use.

Benchmark suites such as GLUE~\cite{wang2018glue}, SuperGLUE~\cite{wang2019superglue}, and MMLU~\cite{hendrycks2021mmlu} have accelerated NLP research by offering standardized testing for a range of tasks. Mental health NLP lacks such benchmark suites. While individual datasets have been shown to support good model performance, the absence of a common resource causes reproducibility, cross-task generalizability, and comparing approaches on an equal footing to be difficult. Several recent studies~\cite{himmi2024towardsrobustNLP} highlight this limitation and call for the development of standardized benchmarks for health-targeted NLP.

In contrast to previous efforts that introduced single-task corpora or collaborative challenges, our contribution unifies four empirically established Reddit-derived datasets into a comprehensive benchmark covering suicidal ideation detection, bipolar disorder detection, general mental health condition detection, and multi-class mental condition classification. Each dataset has been validated in prior work, both empirically and linguistically, with inter-annotator agreement consistently exceeding Cohen's $\kappa=0.8$. By consolidating these resources, we provide a robust and diverse benchmark suite that can serve as a reference for future research in mental health detection, reproducibility studies, and multi-task learning.

\section{Datasets}

Our dataset spans four classification tasks:

\subsection{Suicidal Ideation}

The suicidal ideation data were collected from Reddit for three months (October-December 2022). The suicidal ideation-related subreddit, r/SuicideWatch, and non-suicidal related communities (e.g., r/socialanxiety, r/TrueOffMyChest, r/bipolar, r/confidence, r/geopolitics) were the source of dataset posts. Subreddit selection was based on two considerations: (i) community size and activity, to guarantee sufficient data coverage, and (ii) existence of posts that could reasonably be classified as suicidal or non-suicidal. After this filtering, we collected 37,821 posts, which we proceeded to annotate into the two categories: suicidal and non-suicidal.

\subsubsection{Linguistic Review}
To find typical suicidal communication phrases, we used the TextRank algorithm~\cite{mihalcea-tarau-2004-textrank} through the PyTextRank library\footnote{https://pypi.org/project/pytextrank/}. This picked up highly pertinent expressions such as ``so much pain", ``no hope", ``suicidal thoughts", ``my suicide note", ``no other way", ``just a burden", and ``my own death". These phrases are strongly indicative of suicidal ideation and align with established clinical descriptions of emotional distress. 

\begin{table}[htbp]
    \caption{An overview of linguistic analysis of the suicidal dataset}
    \centering
    \begin{tabular}{|c||cc|}
        \hline
         \textbf{Linguistic} & \multicolumn{2}{|c|}{\textbf{Values}} \\
        \cline{2-3}
         \textbf{Attributes} & \textbf{Suicidal Posts} & \textbf{Non-suicidal Posts} \\
        \hline
        Avg hashtags & 0.034 & 0.048 \\
        \hline
        Avg posts containing URL & 0.002 & 0.022 \\
        \hline
        Avg character length of posts & 851.243 & 1053.305 \\
        \hline
        Avg tokens & 77.022 & 94.040 \\
        \hline
        Avg nouns & 30.855 & 38.539 \\
        \hline
        Avg pronouns & 26.640 & 32.339 \\
        \hline
        Avg verbs & 41.169 & 46.872 \\
        \hline
        Avg Adjectives & 13.364 & 15.812 \\
        \hline
    \end{tabular}
    \label{tab:suicidal_linguistic_analysis}
\end{table}

Table~\ref{tab:suicidal_linguistic_analysis} summarizes key linguistic differences between the groups. Suicidal posts averaged 851 characters and 77 tokens, and contained fewer hashtags and URLs than non-suicidal posts, which were generally longer. Additionally, suicidal messages showed more frequent use of swear words (e.g., f**k, bullsht, btch), reflecting heightened emotional intensity.

\subsubsection{Judgmental Validation}

To evaluate the consistency of our annotations, we conducted a human judgment trial on a random sample of 756 posts (approximately 2\% of the dataset). Two coders independently annotated all posts using a simple rubric: 1) \textbf{Suicidal ideation:} Posts containing an overt intention, discussion of past attempts, or concrete planning (e.g., ``If I can work it out, it's over tonight''). 2) \textbf{Non-suicidal:} Discussion of public matters (e.g., celebrity suicide) or casual conversation without any mention of suicidal ideation (e.g., politics, sports, vacations). 

We measured inter-annotator agreement using Cohen's $\kappa$. The resulting scores, $\kappa = 0.897$ and $\kappa = 0.854$, indicate agreement in all but a few nearly perfect cases. These values not only meet but also surpass the traditional reliability threshold of 0.80~\cite{Landis1977}, demonstrating that our labeling guidelines were reproducible and reliable.

\subsection{Bipolar Disorder}
\label{label:bipolar_dataset}

We collected the bipolar disorder dataset between December 2021 and December 2022. We took posts on \textit{r/bipolar} as examples of the positive class since the subreddit is dedicated to living with bipolar disorder. We took posts from mental illness forums (e.g., \textit{r/depression}, \textit{r/socialanxiety}, \textit{r/TrueOffMyChest}) and also different forums (e.g., \textit{r/confidence}, \textit{r/geopolitics}) as the negative class. This configuration allowed us to create an adequately balanced dataset comprising: (i) hard negative samples from various mental illness states, and (ii) unique samples of other classes. Following conventional practice in mental health NLP~\cite{coppersmith2015adhd}, we applied self-identification patterns to mark the posts where users openly declared their bipolarity. In a bid not to contaminate data, we adhered strictly to the Cohan et al.'s~\cite{cohan2018smhd} filtering guidelines and eliminated duplicate users commenting on bipolar and non-bipolar subreddits. Preprocessed data was approximately 23,000 bipolar posts and 26,000 non-bipolar posts, making it an approximately balanced dataset suitable for application in classification.

\subsubsection{Linguistic Review}

\begin{table}[htbp]
\centering
\caption{Linguistic Differences Between Bipolar and Non-Bipolar Posts}
\label{tab:bipolar_linguistic_features}
\begin{tabular}{|l|c|c|}
\hline
\textbf{Feature} & \textbf{Bipolar} & \textbf{Non-Bipolar} \\
\hline
Average hashtags per post & 0.039 & 0.230 \\
Average URLs per post     & 0.014 & 0.153 \\
Average post length (characters) & 894.36 & 959.17 \\
Average tokens per post   & 80.83 & 93.06 \\
\hline
Average nouns per post    & 33.80 & 42.81 \\
Average verbs per post    & 38.68 & 34.28 \\
Average adjectives per post & 14.76 & 13.78 \\
Average pronouns per post & 24.18 & 19.00 \\
\hline
\end{tabular}
\end{table}

Table~\ref{tab:bipolar_linguistic_features} reveals significant linguistic class distinctions. Non-bipolar posts contained more external references (0.23 hashtags, 0.15 URLs per post) than bipolar posts (0.04 and 0.01), which is evidence of a more outward style. Bipolar posts registered a more expressive and inward style, with higher verb use (38.7 vs. 34.3), higher pronoun use (24.2 vs. 19.0), and slightly more adjectives (14.8 vs. 13.8), consistent with findings by Mariani et al.~\cite{mariani2020linguistic}. Non-bipolar posts contained more nouns (42.8 vs. 33.8), similar to more descriptive, topic-oriented writing~\cite{szabo2024predictive}.

\subsubsection{Sentiment Variance}

\begin{figure}[ht]
    \centering
    \includegraphics[width=0.8\linewidth]{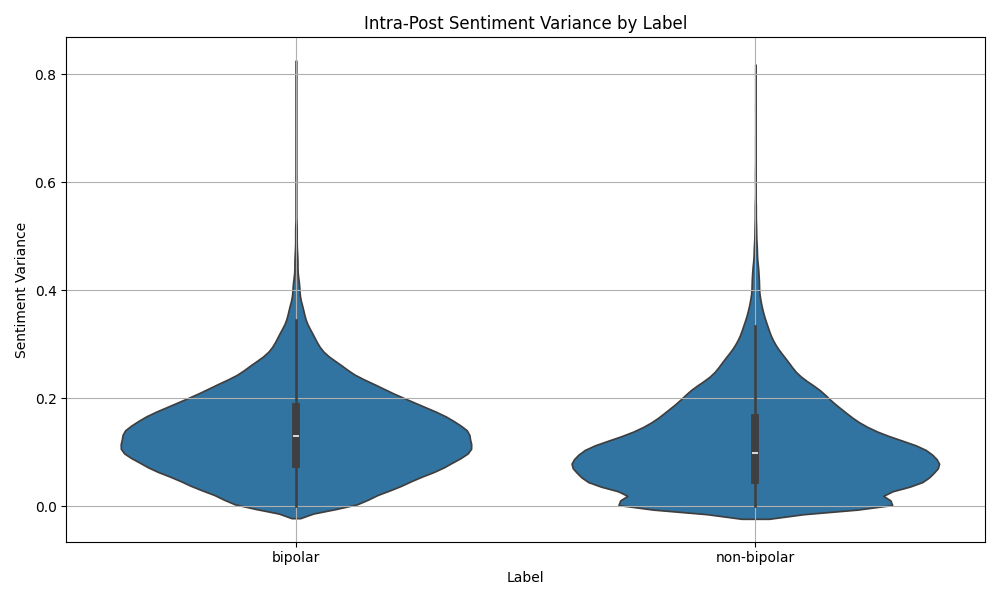}
    \caption{Intra-post sentiment variance distribution for non-bipolar and bipolar posts.}
    \label{fig:sentiment_variance}
\end{figure}

To further distinguish the two groups, we identified intra-post sentiment variance. As it is quite clear from  Figure~\ref{fig:sentiment_variance}, the bipolar posts were of greater variance and greater spread, as could be expected by the emotional oscillation characteristic of bipolar expression. This distinction was highly significant to a Mann–Whitney U test ($U = 373,025,861.5$, $p < 1e^{-280}$).

\subsubsection{Judgmental Validation}

To further ensure annotation quality, we conducted a human evaluation. A stratified sample of 1,006 posts (approximately 2\% of the total data) was independently labeled by two annotators using minimal guidelines: posts were marked as bipolar if they contained signs of extreme mood fluctuations, such as impulsivity, depressive spirals, or rapid shifts in energy levels. All other posts were labeled as non-bipolar.

Inter-annotator agreement was strong, with Cohen’s $\kappa$ values of 0.83 and 0.875—both falling well within the “almost perfect” agreement range according to established benchmarks~\cite{Landis1977}.

These results reinforce that bipolar-related posts are emotionally and linguistically distinct from non-bipolar content. The high inter-rater reliability, supported by both human and machine validation, confirms that the dataset is robust and suitable for use in downstream modeling tasks.

\subsection{General Mental Disorder}
\label{label:general_mental_dataset}

The general mental disorder corpus was constructed from Reddit comments until December 2022. Data were crawled from the mental health-focused subreddits (r/ADHD, r/anxiety, r/bipolar, r/depression, r/CPTSD, r/schizophrenia, r/BPD, r/SocialAnxiety, r/offmychest) and control subreddits that have nothing to do with mental health (r/geopolitics, r/confidence, r/politics, r/bicycletouring, r/sports, r/travel). Subreddits were chosen based on their active user communities and sufficient text content to ensure an equal representation of the mental health and control classes.

Following established procedures~\cite{coppersmith2015adhd}, we used self-identification to identify posts where users disclosed their mental health status. To prevent contamination, we identified users who posted on both control and mental health subreddits, and eliminated their posts according to the approach of Cohan et al.~\cite{cohan2018smhd}. After processing, the dataset had 144,000 labeled posts, split into two sets:

\begin{itemize}
    \item \textbf{Primary set:} 120,000 posts from \textit{r/ADHD}, \textit{r/anxiety}, \textit{r/bipolar}, \textit{r/depression}, \textit{r/CPTSD}, \textit{r/schizophrenia}, and control subreddits (\textit{r/confidence}, \textit{r/politics}, \textit{r/bicycletouring}, \textit{r/sports}). This set was held for training, validation, and hold-out testing.
    
    \item \textbf{External set:} 24,000 posts from \textit{r/BPD}, \textit{r/SocialAnxiety}, \textit{r/offmychest}, \textit{r/travel}, and \textit{r/geopolitics}. This set was held only for external validation, estimating the model's ability to generalize to unseen disorders and control topics.
\end{itemize}

\subsubsection{Linguistic Analysis}

We employed the TextRank algorithm~\cite{mihalcea-tarau-2004-textrank} to identify common expressions associated with mental health issues. Frequent words were: \textit{panic attack, my psychiatrist, my mental health, suicidal thoughts, antipsychotics, severe depression, emotional abuse, childhood trauma, manic episodes}.

\begin{table}[htbp]
    \caption{Linguistic Analysis of Mental Disorder and Control Group Posts}
    \centering
    \begin{tabular}{|c||c|c|}
        \hline
        \textbf{Linguistic Metrics} & \textbf{Mental Disorder} & \textbf{Control Group} \\
        \hline
        Avg. hashtag usage & 0.095 & 0.147 \\
        \hline
        Avg. posts containing URLs & 0.030 & 0.261 \\
        \hline
        Avg. character count per post & 1285.28 & 852.12 \\
        \hline
        Avg. token count per post & 114.03 & 86.59 \\
        \hline
        Avg. nouns & 46.69 & 42.87 \\
        \hline
        Avg. pronouns & 36.25 & 11.57 \\
        \hline
        Avg. verbs & 55.52 & 25.25 \\
        \hline
        Avg. adjectives & 18.68 & 11.32 \\
        \hline
    \end{tabular}
    \label{tab:general_mental_linguistic_analysis}
\end{table}

Table~\ref{tab:general_mental_linguistic_analysis} presents language differences between mental health and control posts:
\begin{itemize}
    \item \textbf{Hashtags and URLs:} Control group posts included more hashtags (0.147 vs. 0.095) 
    and URLs (26.1\% vs. 3.0\%), indicating more extroverted, information-like arguments.
    
    \item \textbf{Post length:} Mental health posts were significantly longer (1285 vs. 852 characters; 114 vs. 87 tokens), reflecting more information-reporting stories and more elaborate personal anecdotes.
    
    \item \textbf{POS distribution:} Mental health posts contained a greater share of pronouns (36.3 vs. 11.6), verbs (55.5 vs. 25.3), and adjectives (18.7 vs. 11.3) in usage, which corresponds to personal anecdotal and emotionally expressive language.
\end{itemize}

The findings are that mental health posters post longer, more personal, and more affective messages, and control group messages are shorter, fact-based, and more other-oriented.

\subsubsection{Judgmental Validation}

To test for annotator consistency, we conducted a human judgment test on a stratified sample of 1,000 postings. Two annotators labeled postings based on a short rubric:

\begin{itemize}
    \item \textbf{Mental Disorder:} posts explicitly mentioning mental illness, emotional disturbance, or a clinical diagnosis.
    \item \textbf{Control:} posts on general topics unrelated to mental health (e.g., hobbies, politics, sports) or one's mental illness.
\end{itemize}

Quantification of agreement was performed using Cohen's $\kappa$~\cite{Cohen1960}. Both annotators achieved $\kappa$ scores above 0.94, which falls in the ``almost perfect" category~\cite{Landis1977}. These results provide strong evidence that the dataset is reliable, reproducible, and minimally affected by subjective bias, making it a credible foundation for downstream general mental disorder detection.

\subsection{Multi-class Mental Disorder}
\label{label:multi_class_dataset}

We constructed a multi-class mental disorder dataset using Reddit posts with a cutoff date of December 2022. The dataset draws from six subreddits: \textit{r/ADHD}, \textit{r/Anxiety}, \textit{r/Bipolar}, \textit{r/Depression}, \textit{r/CPTSD}, and \textit{r/Schizophrenia}. These forums are spaces where individuals share personal experiences directly related to the corresponding disorders. Therefore, a post within a specific subreddit is treated as a potential instance of that disorder. 
To build an equally balanced control class, we used threads from popular subreddits with low probabilities of containing mental health discussions like \textit{r/Politics}, \textit{r/BicycleTouring}, \textit{r/Confidence}, and \textit{r/Sports}. Multidomain diversification prevented subject matter confounds and ensured that classification would pick up on linguistic cues of mental health instead of subreddit topics.
 
The initial dataset had over 3.3 million posts across the six disorder classes and control class. We randomly sampled 105,000 posts with an equal number of posts per the seven classes based on the same self-reporting and overlap-user filtering parameters described in Section~\ref{label:general_mental_dataset}. This dataset can be used as the core foundation for multi-class model training and testing. 

\subsubsection{Linguistic Review}

\begin{table}[htbp]
    \vspace{-2mm}
    \caption{Key Stylistic and POS Differences Across Classes}
    \label{tab:multi_class_mental_linguistic_analysis}
    \centering
    \small
    \begin{tabular}{lrrrrr}
        \hline
        \textbf{Class} &
        \begin{tabular}[c]{@{}c@{}}\textbf{URLs}\\[-2pt](avg)\end{tabular} &
        \begin{tabular}[c]{@{}c@{}}\textbf{Chars}\\[-2pt](avg)\end{tabular} &
        \begin{tabular}[c]{@{}c@{}}\textbf{Tokens}\\[-2pt](avg)\end{tabular} &
        \begin{tabular}[c]{@{}c@{}}\textbf{Nouns}\\[-2pt](avg)\end{tabular} &
        \begin{tabular}[c]{@{}c@{}}\textbf{Verbs}\\[-2pt](avg)\end{tabular} \\ 
        \hline
        Control        & 0.14 & 794.4 & 79.6 & 39.5 & 25.9 \\
        ADHD           & 0.04 & 1178.4 & 106.3 & 45.1 & 49.9 \\
        Anxiety        & 0.01 & 1243.5 & 111.8 & 45.6 & 54.1 \\
        Bipolar        & 0.02 & 967.9 & 86.9 & 35.6 & 41.7 \\
        CPTSD          & 0.04 & 1840.2 & 160.3 & 65.8 & 78.2 \\
        Depression     & 0.01 & 1523.97 & 133.5 & 52.6 & 68.9 \\
        Schizophrenia  & 0.07 & 1062.1 & 93.6 & 38.9 & 44.9 \\
        \hline
    \end{tabular}
    \vspace{-2mm}
\end{table}

Table~\ref{tab:multi_class_mental_linguistic_analysis} shows stylistic distinction by category. Control posts were brief ($\approx$80 tokens, 800 characters), and CPTSD posts were roughly twice as long ($\approx$160 tokens, 1.8K characters). Verb usage followed similarly, between an estimated 26 in Control posts and 78 in CPTSD posts, and may be one indication of more lengthy, text-dense narrative text. Noun usage increased in disorder content as well.

Control posts had greater external linking (0.14 per post), whereas posts relating to disorder had lower levels of linking ($\leq$0.07) with a recommendation to use personal issues over external.

\begin{figure}[htbp]
    \centering
    \includegraphics[width=0.6\linewidth]{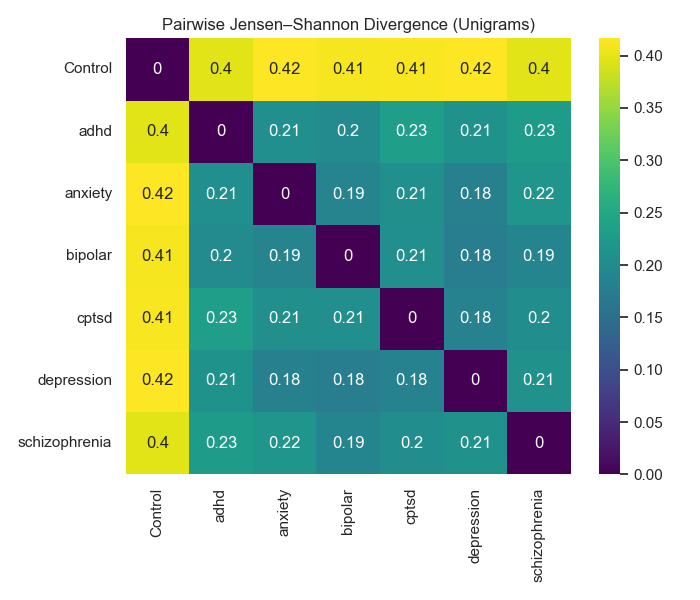}
    \caption{Pairwise divergence map across classes}
    \label{fig:divergence_heat_map}
\end{figure}

Figure~\ref{fig:divergence_heat_map} shows pairwise Jensen–Shannon divergence between unigram language models for all the classes~\cite{lin1991divergence}. Jensen–Shannon divergence ranges from 0 (indicating total agreement) to 1 (representing maximum lexical separation). Disorder classes were lexically farthest from Control posts (mean $JS \approx 0.41$), but nearest among disorders ($JS \approx 0.18$–$0.23$). Most similar was Depression and Bipolar ($JS = 0.18$), seen through shared affective lexicon. These findings justify that every class has a distinctive linguistic signature, justifying the quality of annotations for downstream modeling.

\subsubsection{Judgmental Validation}

As an additional measure of label quality, we reserved 1,050 posts (approximately 1\% of data) for human judgment. Two independent annotators labeled posts according to the following rubric:

\begin{itemize}
    \item \textbf{Disorder classes (ADHD, Anxiety, Bipolar, CPTSD, Depression, Schizophrenia):} explicit self-disclosure of diagnosis, description of symptoms (e.g., panic attacks, manic episodes, hallucinations), mention of medication, or a call for treatment.
    
    \item \textbf{Control:} posts with no self-identified mental health content, typically discussing everyday, political, or recreational activities without clear emotional distress.
\end{itemize}

Agreement was measured using Cohen's $\kappa$; scores of 0.92 and 0.91 place the results in the ``almost perfect'' agreement category~\cite{Landis1977}. These findings indicate that the annotation guidelines are precise, reproducible, and sufficient to train consistent multi-class classifiers.

\section{Performance and Reliability}

To establish the reliability of our datasets as research resources, we present a summary of key performance benchmarks from our earlier work, published across four independent studies~\cite{hasan2024suicidal}~\cite{hasan2025bipolar}~\cite{hasan2025generalmental}~\cite{hasan2025multiMental}. In these studies, each dataset was evaluated on a specific classification task, ranging from suicidal ideation detection to multi-class disorder classification. Representative results obtained with state-of-the-art models are presented in Table~\ref{tab:performance_benchmarks}. Our findings show that transformer-based models such as RoBERTa performed well across the board, while contextually embedded recurrent models (e.g., LSTMs using BERT embeddings) also achieved strong, competitive performance.

\begin{table}[htbp]
    \caption{Performance Benchmarks across the Four Datasets}
    \centering
    \begin{tabularx}{\columnwidth}{|X||X||X|}
        \hline
        \textbf{Dataset} & \textbf{Representative Models} & \textbf{Reported Performance} \\
        \hline
        Suicidal Ideation Detection & RoBERTa, BERT+LSTM & F1 up to 93.14\% \\
        \hline
        Bipolar Disorder Detection & RoBERTa, BERT+LSTM & F1$\approx$98\% \\
        \hline
        General Mental Disorder (Binary) & RoBERTa, BERT+LSTM & F1 up to 99.54\% (hold-out), 95.96\% (external) \\
        \hline
        Multi-Class Mental Disorder Classification & DistilBERT, BERT+BiLSTM+Attn & F1 ranges 88.03\%--99.20\% \\
        \hline
    \end{tabularx}
    \label{tab:performance_benchmarks}
\end{table}

Across all tasks, transformer models such as RoBERTa achieved uniformly high F1 scores (93–99\%), and LSTMs enhanced with contextual embeddings achieved competitive results. This pattern supports the fact that the datasets provide consistent discriminative signals, enabling subsequent fine-tuned models to reproduce strong results. 
Random samples of all four datasets were also human-annotated with meticulously designed guidelines. Inter-annotator agreement was consistently high (Cohen's $\kappa > 0.8$), falling within the ``almost perfect'' range~\cite{Landis1977}. These measures confirm that the datasets are not only learnable but also reliable as gold-standard references.

Furthermore, the datasets cover a spectrum of tasks—including binary detection (general mental disorder), disorder-specific detection (bipolar disorder, suicidal ideation), and multi-class classification—forming a comprehensive benchmark suite. This variety enables researchers to evaluate models across diverse challenges, from broad binary tasks to fine-grained classification. Overall, the performance results and annotation quality demonstrate that the datasets are consistent, learnable, and suitable as benchmarks for future research. 

\section{Conclusion}

This paper consolidates four previously developed Reddit-based datasets—focused on suicidal ideation, bipolar disorder, general mental disorder, and multi-class mental disorder—into a unified benchmark resource for NLP in mental health. While earlier work treated these datasets separately, this study highlights their collective value by integrating linguistic insights, annotation protocols, and performance evaluations into a single framework.

The primary contribution of this work lies in elevating these datasets from isolated task-specific tools to a reproducible, diverse, and methodologically consistent benchmark suite. This integration opens two promising directions for future work:

\begin{itemize}
\item \textbf{Multi-task learning:} The complementary nature of these datasets can be harnessed to train models that identify common linguistic and psychological markers across different mental health conditions.
\item \textbf{Benchmark standardization:} By offering a unified evaluation framework, this suite helps reduce methodological fragmentation in mental health NLP and supports more reproducible and comparable research.
\end{itemize}

In summary, this work reframes four distinct datasets as a cohesive, validated, and reusable resource designed to accelerate progress in mental health NLP—enabling richer multi-task models and more standardized benchmarking in the field.


%
%
\bibliographystyle{splncs04}
\bibliography{references}

\end{document}